# Real-time Deer Detection and Warning in Connected Vehicles via Thermal Sensing and Deep Learning

**Hemanth Puppala**
Holcombe Department of Computer and Electrical Engineering
Clemson University, Clemson, SC  29634
Email:  hpuppal@clemson.edu

**\*Wayne Sarasua, Ph.D., P.E.**
Glenn Department of Civil Engineering
Clemson University, Clemson, SC  29634
Email:  sarasua@clemson.edu

**Srinivas Biyaguda**
Clemson School of Computing
Clemson University, Clemson, SC  29634
Email:  sbiyagu@clemson.edu

**Farhad Farzinpour**
Glenn Department of Civil Engineering
Clemson University, Clemson, SC  29634
Email:  ffarzin@clemson.edu

**Mashrur Chowdhury, Ph.D., P.E.**
Glenn Department of Civil Engineering
Clemson University, Clemson, SC  29634
Email:  mac@clemson.edu

Word Count: 6356 words + 4 table (250 words per table) = 7356 words

*Submitted 8/1/2025*
\*Corresponding Author

**ABSTRACT**
Deer-vehicle collisions represent a critical safety challenge in the United States, causing nearly 2.1 million incidents annually and resulting in approximately 440 fatalities, 59,000 injuries, and $10 billion in economic damages. These collisions also contribute significantly to declining deer populations. This paper presents a real-time detection and driver warning system that integrates thermal imaging, deep learning, and vehicle-to-everything communication to help mitigate deer-vehicle collisions. Our system was trained and validated on a custom dataset of over 12,000 thermal deer images collected in Mars Hill, North Carolina. Experimental evaluation demonstrates exceptional performance with 98.84% mean average precision, 95.44% precision, and 95.96% recall. The system was field tested during a follow-up visit to Mars Hill and readily sensed deer providing the driver with advanced warning. Field testing validates robust operation across diverse weather conditions, with thermal imaging maintaining 88-92% detection accuracy in challenging scenarios where conventional visible-light based cameras achieve less than 60% effectiveness. When a high probability threshold is reached sensor data sharing messages are broadcast to surrounding vehicles and/or roadside units via cellular vehicle to everything (C-V2X) communication devices. Overall, our system achieves end-to-end latency consistently under 100ms from detection to driver alert. This research establishes a viable technological pathway for reducing deer-vehicle collisions through thermal imaging and connected vehicles. While the system evaluation provides proof of concept, a primary limitation of the system is the detection range due to resolution. Further testing with higher resolution cameras will improve the system's utility.

**INTRODUCTION**

Deer-vehicle collisions (DVCs) are among the most significant wildlife-related transportation safety challenges in North America. Recent statistics show that U.S. drivers face a 1 in 128 chance of striking an animal each year, with over 1.8 million animal collision insurance claims filed between July 2023 and June 2024 *(1)*. Deer accounts for approximately 1.3 million of these incidents, making up about 72% of all animal-vehicle collision insurance claims during that period *(2)*. It is estimated that more than 2.1 million DVCs occur annually, resulting in approximately 440 human fatalities and nearly 60,000 injuries, highlogting the serious threat to public safety *(1)*.

The economic burden extends far beyond immediate vehicle damage costs. While direct property damage claims average over $4,300 per incident, the total annual societal cost exceeds $10 billion when accounting for medical expenses, emergency response services, traffic disruption, and lost productivity (1). In California alone, wildlife-vehicle collisions cost more than $200 million annually, with deer representing the most frequently struck large animal (3).

The consequences of vehicle collisions extend significantly beyond human casualties to wildlife conservation. Vehicle strikes have emerged as a leading cause of deer population decline across multiple regions. The University of California Davis Road Ecology Center's 2024 report reveals that more than 48,000 deer are killed annually by vehicles on California roads alone—more than twice the number harvested by hunting. These collisions account for over 10% of deer deaths annually in the state, where the total deer population has been declining for decades (3).

Deer-vehicle collisions exhibit distinct temporal and geographic patterns that inform mitigation strategies. The vast majority of incidents occur during the twilight hours of dawn and dusk, and during nighttime hours, coinciding with peak deer activity and reduced driver visibility *(1)*. October through December represents the highest-risk season, with peak collision rates occurring during the autumn mating season when deer movement increases substantially. Pennsylvania leads in absolute collision numbers and generating the highest volume of insurance claims nationwide *(1, 2)*.

Conventional mitigation strategies have demonstrated limited scalability and effectiveness. Passive measures such as static warning signs often lead to driver habituation, showing minimal impact on collision reduction. While properly maintained fencing paired with wildlife crossings can reduce DVCs by over 80%, implementation costs limit application to short, high-priority corridors, leaving the majority of at-risk road networks unprotected *(4)*. The temporal concentration of collisions during low-visibility periods highlights a critical limitation of existing approaches. Traditional visible-light detection systems fail when needed most—during darkness when deer's are most active, and collision risk is highest *(5)*.

Recent studies have emphasized the potential of infrared (IR) sensors for real-time hazard detection, highlighting their effectiveness in low-light conditions, rapid response times, and low energy consumption, which make them suitable for highway safety applications *(29)*. These findings motivate the use of thermal infrared imaging as an advanced modality. Unlike conventional IR sensors that primarily detect motion or beam interruptions, thermal cameras capture long-wave infrared radiation, enabling robust detection of living beings such as deer under diverse lighting and weather conditions.This research addresses the limitations through an innovative integration of thermal imaging, edge artificial intelligence, and Vehicle-to-Everything (V2X) communication technologies. The proposed system leverages thermal cameras' superior performance in challenging visibility conditions, combining real-time deep learning inference with collaborative safety messaging via standardized V2X protocols. The driver of the sensing vehicle is warned instantaneously as a deer is sensed. The V2X connected vehicle technology warns other drivers as they approach the area where deer were initially detected.

The versatility of this high accuracy, low-latency perception system extends beyond driver warnings to enable a wide range of advanced active safety features. For instance, it can serve as a critical front-end technology for sophisticated collision avoidance systems like Evasive Minimum Risk Maneuvering (EMRM), which are designed to execute aggressive maneuvers to mitigate severe losses in high-risk edge cases. The effectiveness of such advanced maneuvers is critically dependent on early and accurate hazard detection, requiring high-precision perception modules to provide the "timely and accurate information" necessary for preemptive action. More broadly, this principle applies to the entire suite of Advanced Driver Assistance Systems (ADAS), for which high-resolution, real-time perception data are considered a critical foundation for functions like crash prediction and driver behavior modeling. Our system, with its sub-100ms end-to-end latency and 98.84% mAP, is specifically engineered to meet these stringent requirements. By leveraging thermal imaging, it provides robust environmental awareness in the challenging low-light conditions where conventional visible-light systems often fail, addressing a significant gap in current ADAS sensor capabilities. Furthermore, the planned expansion to multi-class detection, including other wildlife and pedestrians, positions this technology not just as a solution for DVCs, but as a versatile perception module that can be integrated into a comprehensive ADAS sensor suite to mitigate a wide range of collision scenarios *(30,31)*

The paper discusses previous work, the system architecture, training and sensor development, field testing and experimental results, limitations, and future research.

## LITERATURE REVIEW

The literature review focused on animal detection using thermal cameras and detection algorithms, and also provides a short discussion on V2X communication. It is noteworthy that there has been a great deal of research done with pedestrian detection *(4, 6-8)*. Animal detection presents unique challenges compared to pedestrian detection, primarily because animals are often partially covered by vegetation and display a much wider range of appearances depending on the direction from which they are viewed. This variability makes it significantly harder for vision systems to consistently recognize animals in natural environments than pedestrians in urban ones.

**Thermal Imaging Technology Evolution**
Thermal imaging technology has undergone significant advancement, becoming increasingly accessible and effective for wildlife detection applications. Unlike visible-light systems that depend on reflected photons, thermal cameras detect infrared radiation emitted by objects, enabling clear visibility regardless of ambient lighting conditions *(9)*. Modern thermal sensors no longer require bulky cooling systems and have achieved substantial cost reductions while maintaining high sensitivity.

The technology offers distinct advantages for wildlife detection: superior performance in low-light scenarios, immunity to visual camouflage, capability to detect heat signatures through moderate vegetation, and reduced false alarms compared to motion-based systems. These characteristics prove particularly valuable for deer detection, as deer body heat creates distinctive thermal signatures against cooler environmental backgrounds.

**Previous Thermal Detection Research**
Early research by Zhou *(10)* demonstrated thermal-based deer detection achieving up to 100% accuracy at 10 Hz processing speed using a contour-based histogram of oriented gradients algorithm. The system was mounted on a roadside lightpole and successfully detected deer across various weather conditions, establishing the viability of thermal imaging for DVC mitigation.

Forslund and Bjarkefur *(11)* discuss a thermal night vision animal detection system that was previously installed in european automobiles. It uses a classical boost classifier to detect animals. Classified detections of an animal are linked together into an animal track that is used to determine if a warning message should be generated. Detection accuracy is improved as a vehicle approaches the detected animal. The authors indicted that false positives can be virtually eliminated if an animal is tracked long enough.

Recent advances in deep learning have substantially improved thermal detection capabilities. Research by Lyu et al. used a modified faster region-based convolutional neural network (R-CNN) for thermal deer detection from unmanned aerial vehicle (UAV) imagery, achieving 92.3% mean Average

Precision (mAP) for all objects, 78.9% for small objects, 94.6% for medium objects, and 95.8% for large objects *(12)*.

Comparative studies reveal significant performance advantages of thermal systems over visible-light alternatives. Research demonstrates that under adverse weather conditions, thermal models achieved 97.9% mean Average Precision (mAP) while RGB models achieved only 19.6% mAP *(13)*. One detection probability study comparing thermal imaging with traditional spotlight methods found thermal systems detecting 92.3% of deer compared to 54.4% for spotlight surveys *(5)*.

**Deep Learning Architecture Development**
The You Only Look Once (YOLO) framework has provided significant advances in real-time object detection by reformulating detection as a single regression problem, enabling processing speeds necessary for automotive applications *(14)*. YOLOv8, which is used in this research, offers improved accuracy and efficiency over previous versions through enhanced feature extraction and optimized anchor-free detection heads.

Recent research demonstrates YOLO's effectiveness for thermal applications. Enhanced YOLOv8 variants optimized for thermal imagery have achieved up to 92.3% mAP for small object detection *(9)*. The single-stage architecture provides speed advantages crucial for real-time applications, while architectural improvements in feature extraction and multi-scale detection enhance performance on small and distant objects—critical capabilities for wildlife detection scenarios.

**V2X Communication Standards and Implementation**
The National ITS Reference Architecture provides a comprehensive list of potential connected vehicle service packages or applications that can enable V2X (Vehicle-to-Everything) capabilities (https://www.arc-it.net/). These applications cover a broad range of use cases, including vehicle safety, public safety, and traffic management. Each of these applications plays a critical role in enhancing road safety and improving the operational efficiency of surface transportation systems.

To support seamless communication among the various components involved in these applications, V2X standards have been developed. These standards ensure reliable data exchange between different physical elements of the connected vehicle ecosystem, such as vehicles, roadside units, traffic management centers, and emergency response systems.

V2X communication encompasses multiple standards developed for automotive safety applications. The SAE J2735 standard defines the V2X Communications Message Set Dictionary, specifying data structures and message formats for vehicle-to-vehicle, vehicle-to-infrastructure, and vehicle-to-device communications *(15)*. This standard includes Basic Safety Messages (BSMs) for general vehicle status and specialized messages like Sensor Data Sharing Messages (SDSMs) for dynamic environmental hazard alerts.

SAE J2945 series standards define performance requirements and implementation guidelines for V2X message usage *(16)*. These specifications ensure interoperability between vehicles and infrastructure from different manufacturers, forming the foundation for collaborative safety systems. The message formats use Abstract Syntax Notation One (ASN.1) encoding to support broad compatibility across communication media *(15)*.

A study by Enan et al. developed a vision-based safety application for pedestrian safety using V2X(communication technologies. In this application, if any risky pedestrian behavior is detected, the system sends warnings to the pedestrians and vehicles exhibiting such behavior. The authors presented a video analytics-based Basic Safety Message (BSM) generation method that complies with the Society of Automotive Engineers (SAE) standard, specifically SAE J2945 and SAE J2735 *(16B)*.

Contemporary V2X deployments support dual-mode operation combining Dedicated Short-Range Communications (DSRC) and Cellular V2X (C-V2X) technologies, providing communication redundancy and extended range capabilities *(17)*. These systems achieve communication ranges exceeding 1000 meters under optimal conditions while maintaining sub-10ms latency for adjacent vehicle messaging *(18)*.

**SYSTEM DESIGN, WORKFLOW, AND METHODOLOGY**

This section discusses the design of the system including the components and processes that take place. The workflow shown in **Figure 1** integrates hardware and software components with clear data flow and decision pathways. The process begins with thermal image capture, followed by real-time preprocessing and deep learning model inference. When deer detection occurs, the system evaluates thresholds and then sends a warning message to the driver if the driver threshold is met. If the broadcast threshold is met, the system generates a Sensor Data Sharing Message (SDSM) transmitted via User Datagram Protocol (UDP) to connected On Board Units (OBUs). After ASN1 encoding, warning messages are broadcast over-the-air to nearby vehicles and Roadside Units (RSUs) through V2X communication. Continuous monitoring enables ongoing detection cycles, demonstrating the integration of sensing, edge processing, and cooperative safety intervention within a unified system. The following sections discuss the individual system components in detail.

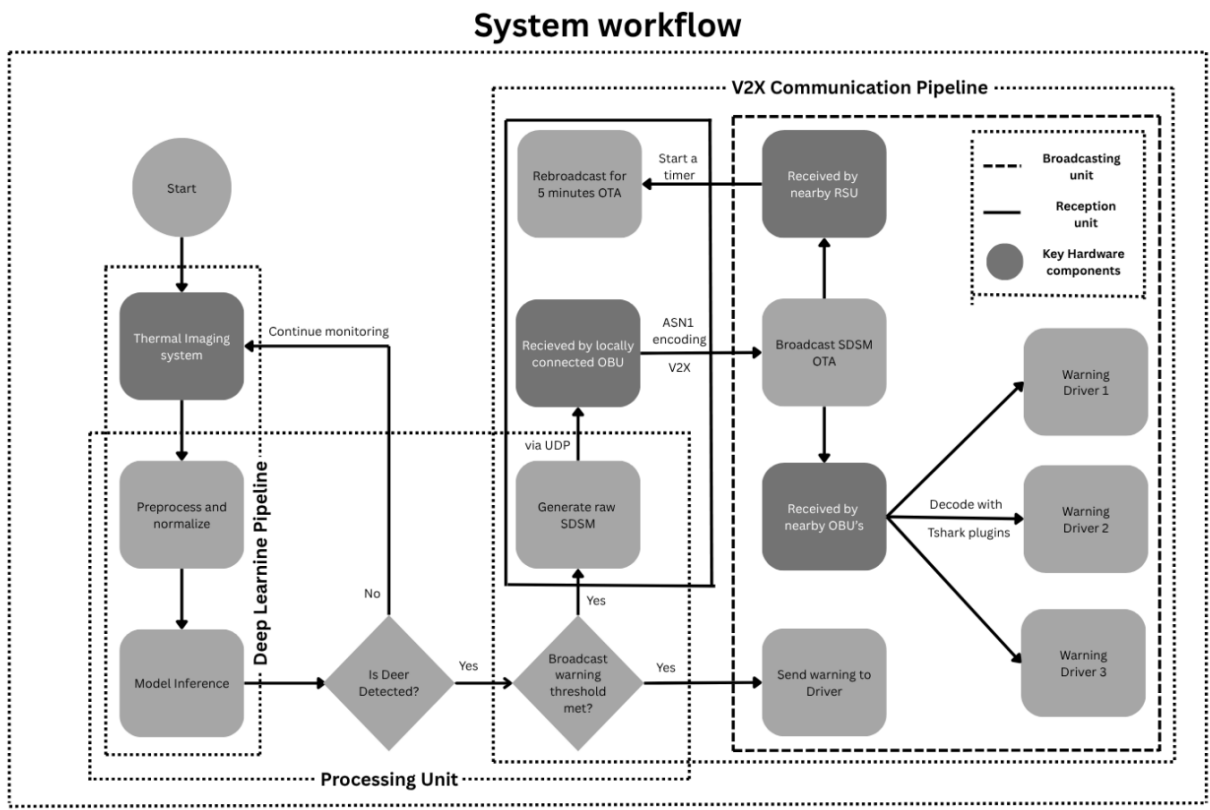

**Figure 1 System workflow for real-time deer detection and V2X alerting**

**Hardware Configuration**

The integrated system comprises three core hardware components optimized for automotive deployment: 1. Processing unit; 2. Thermal imaging system; and 3. V2X Communication devices. There are additional peripheral devices including a computer monitor and a modified sine wave 400-watt power invertor that plugs into a vehicle's cigarette lighter and provides 120 AC power to all of the hardware.

*Processing Unit*

An NVIDIA Jetson Orin Nano Developer Kit configured in 15W high-performance mode serves as the primary computational platform. This embedded system provides sufficient GPU acceleration for real-time YOLOv8s inference while maintaining power consumption suitable for vehicular applications. The platform supports up to 40 TOPS (Tera Operations Per Second) of AI performance with 8GB unified memory.

*Thermal Imaging System*
A TOPDON TC001 USB-C Thermal Imaging Camera with 256×192 resolution serves as the primary sensor hardware for infrared data acquisition. The camera captures thermal signatures that enable deer detection regardless of ambient lighting conditions, addressing the fundamental limitation of visible-light camera systems during dawn, dusk, and nighttime periods when wildlife collisions most frequently occur. **Figure 2** shows the camera mounted to a vehicle's windshield wiper for testing purposes. The camera must be mounted externally because thermal sensors cannot effectively operate through conventional automotive glazing. The camera is designed to work with an android phone. The researchers were able to download and modify python code that allows it to work with the NVIDIA Jetson Orin Nano.

*V2X Communications*
Cohda Wireless MK6 OBUs and RSUs provide comprehensive V2X connectivity, including DSRC, C-V2X, and 5G NR with fallback support for lower cellular technologies. These devices achieve communication ranges exceeding 1,000 meters and offer multi-constellation GNSS positioning. The power requirement for the MK6 OBU is approximately 25W at 7–15V DC input. *(19-20)*

     RSUs provide the infrastructure-based V2X communication hardware that extends system coverage beyond direct vehicle-to-vehicle range limitations. These roadside hardware components receive, process, and rebroadcast warning messages to ensure comprehensive coverage across the communication network, particularly important in rural deployment scenarios where vehicle spacing may exceed direct communication ranges.

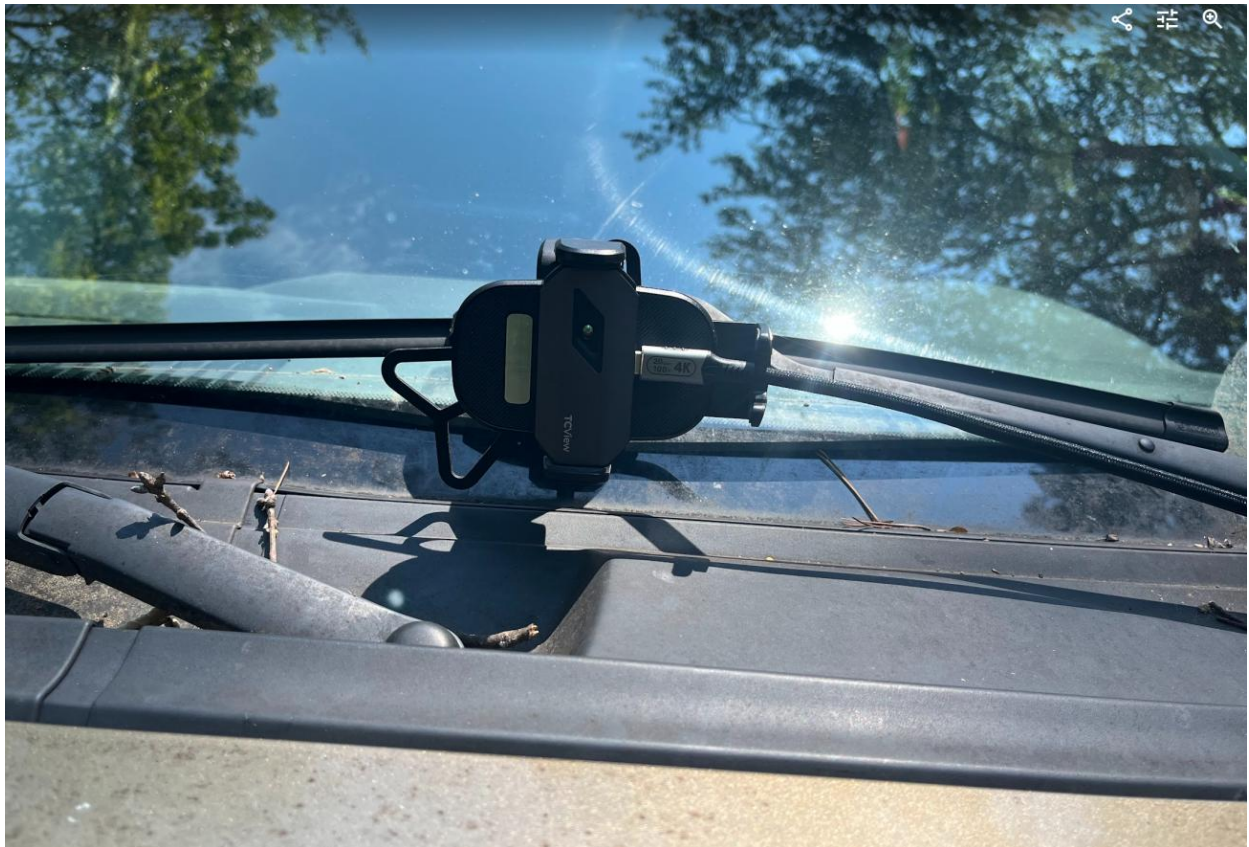
**Figure 2 TOPDON TC001 thermal camera mounted to a vehicle windshield wiper for field testing**

**Software Configuration and Processing Pipeline**
Figure 1 identifies three distinct software processing pipelines that operate within the hardware infrastructure: 1. Deep Learning Pipeline; 2. Processing Unit Pipeline; 3. V2X Communication Pipeline.

*Deep Learning Pipeline*
The deep learning pipeline, enclosed within the dotted boundary on the left side of the workflow, encompasses the complete computer vision and machine learning workflow. This pipeline begins with thermal image preprocessing and normalization algorithms that extract and condition sensor data from raw camera frames, followed by resizing to 256×256-pixel format. The processed images are then fed into model inference using a fine-tuned YOLOv8s model that performs real-time deer detection, providing class probabilities, bounding box coordinates, and detection confidence scores optimized for single-class deer detection.

*Processing Unit Pipeline*
The processing unit pipeline, shown in the center dotted boundary of the workflow, serves as the central control and decision-making layer that bridges detection results with communication actions. Following model inference, this pipeline employs decision-making algorithms that evaluate detection results through the "Is Deer Detected?" decision point and the "Broadcast warning threshold met?" criteria. This processing unit manages the control flow logic and threshold evaluation that determines when detection events warrant driver warnings and V2X message broadcasting.

*V2X Communication Pipeline*
The V2X communication pipeline, enclosed within the dotted boundary on the right side of the workflow, manages the complete messaging workflow from message creation to alert delivery using Cohda's 1609 application framework for V2X message transmission and reception *(21)*. When broadcast thresholds are satisfied, this pipeline generates raw SDSM formatted according to SAE J2735 specifications, incorporating key information elements including object type and identification, measurement timestamps, position offsets, speed measurements, and heading data for each parameter. The pipeline handles ASN1 encoding for V2X standardization, manages UDP transmission to locally connected OBUs, and coordinates over-the-air broadcasting to nearby vehicles and RSUs. Received messages undergo software-based decoding using Tshark with specialized Cohda V2X plugins, enabling extraction of object type, unique identifiers, location coordinates, and metadata for driver alert generation.

**Data Collection and Dataset Development**
Data collection occurred in the Wolf Laurel neighborhood of Mars Hill, North Carolina, near the end of April 2025. The neighborhood was selected for its substantial deer population and diverse environmental conditions. The location provided realistic scenarios encompassing various environmental conditions (e.g. daytime, nighttime, fog, and rain), terrain and vegetation, and variable sight distance. The low traffic volumes allowed the researchers to drive relatively slowly through the neighborhood. The researchers scheduled data collection when deer are most active during twilight hours. In the evenings the data collection began about an hour before sunset and would continue until at least an hour after sunset depending on deer activity. The process was repeated in the morning at dawn. Initial capture yielded approximately 40,000 thermal images collected over 4 days across diverse environmental conditions including clear weather, heavy rain, and dense fog. **Figure 3** shows some sample thermal images. Temperatures ranged from 45 to 65 degrees Fahrenheit.

*Data Curation and Annotation*
From the initial dataset, 12,037 images were selected for annotation. The curation process eliminated blurred, low-quality, and redundant images while maintaining diversity in deer poses, group sizes, distances, and environmental backgrounds. Annotation used the Roboflow platform for single-class deer detection with YOLO-format bounding box annotations *(22)*. The deer shown in sample images in figure 3 are annotated with bounding boxes used for model training and also incorporating preprocessed augmentations to demonstrate the diverse variations applied for improving detection robustness.

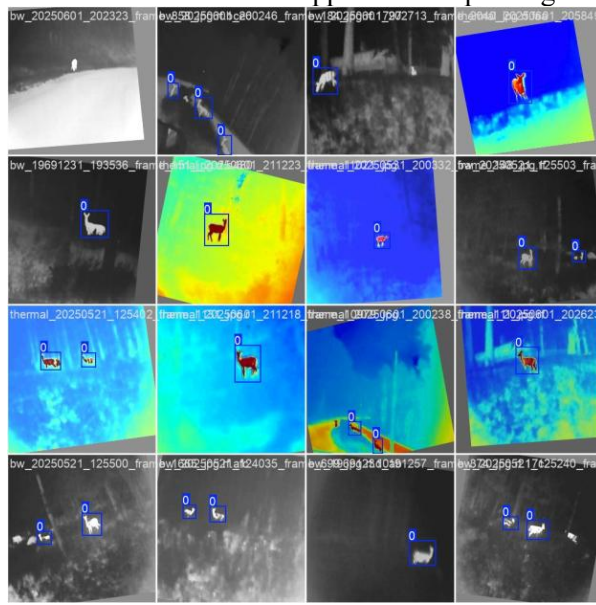

**Figure 3 Annotated deer images with bounding boxes**

## DATASET DEVELOPMENT AND TRAINING STRATEGY

The training dataset implements a strategic 70/30 composition ratio between grayscale thermal images and RGB heatmap representations. This dual-format approach provides several technical advantages: grayscale images preserve raw thermal intensity data crucial for accurate heat signature detection, while RGB heatmaps enhance visual feature contrast through colormap transformations, improving model generalization across different thermal camera calibrations and environmental conditions.

The technical rationale for 70% grayscale images is to retain a significant amount of direct sensor data, preserving true pixel-level thermal values essential for temperature-based object identification. This ensures the model learns fundamental thermal contrast patterns between deer and surrounding environments. The RGB heatmaps (30% of dataset) amplify subtle spatial gradients and edge information through color contrast, functioning as advanced data augmentation while simulating variations in thermal camera output formats encountered in diverse deployment scenarios.

The 12,037 annotated images were divided into training images, validation images, and test images as shown in **Table 1**. The annotated images capture deer at distances ranging from 10-150 feet, with 70% falling within the 10-60 feet range to capture deer features at higher resolution. Extended range samples (60-150 feet) enable earlier warning capabilities while accounting for thermal camera resolution limitations at distance.

The dataset includes diverse object density scenarios with 1-4 deer per frame, though single-deer instances predominate. This distribution reflects real-world encounter patterns while providing sufficient multi-deer examples for robust detection performance across various group configurations.

**Table 1 Dataset Statistics and Characteristics**

| Dataset Component | Count | Percentage | Purpose |
|---|---|---|---|
| Training Images | 9,118 | 75.8% | Model parameter optimization |
| Validation Images | 2,009 | 16.7% | Hyperparameter tuning and monitoring |
| Test Images | 910 | 7.6% | Final performance evaluation |
| **Total Images** | **12,037** | **100%** | **Complete annotated dataset** |

### Training Infrastructure and Configuration

Model training used Clemson University's Palmetto High-Performance Computing cluster, comprising over 1,138 compute nodes with 46,000+ CPU cores and 700+ GPU-equipped systems *(23)*. The recent Palmetto 2 upgrade includes 56 NVIDIA H100 80G GPUs specifically optimized for artificial intelligence/machine learning workloads, providing necessary computational power for large-scale deep learning experiments.

YOLOv8s model training employed transfer learning from pre-trained weights with optimized hyperparameters: 256×256 pixel input resolution matching thermal camera output, batch size 32 for efficient GPU utilization, 180 total epochs with early stopping patience 25, stochastic gradient descent (SGD) optimizer with initial learning rate 0.01 and final rate 0.01, momentum 0.937 with 3-epoch warmup period.

The comprehensive augmentation pipeline included hue (H), saturation (S), value (V) adjustments (H=0.015, S=0.3, V=0.4) optimized for thermal imagery characteristics, geometric transformations including rotation (±15°), translation (10%), and scaling (20%), plus advanced techniques including Mosaic (0.8), MixUp (0.1), perspective distortion (0.0001), and horizontal flipping (0.5).

## EXPERIMENTAL RESULTS AND PERFORMANCE ANALYSIS

**Table 2** summarizes the training and evaluation results for the fine-tuned YOLOv8s model on the deer dataset. The model achieved exceptional detection performance, with a final mAP@0.5 of 98.84%, substantially outperforming baseline YOLOv8s architectures and previous thermal object detection studies *(24)*. The precision, recall, and F1 scores exceeded 95%, indicating both high accuracy and balanced detection.

**Figure 4** presents the training and validation loss curves, as well as key detection metrics, across all epochs. The fine-tuned YOLOv8s model demonstrated excellent convergence over 180 epochs, with a total training time of 4,190 seconds (69.8 minutes) and an average epoch duration of 23.3 seconds. The box regression loss improved by 45%, decreasing from 1.369 to 0.756, while the classification loss reduced by 67% from 0.898 to 0.299 (Figure 3a–d). Validation loss curves closely followed the training loss, exhibiting no signs of overfitting and confirming robust generalization.

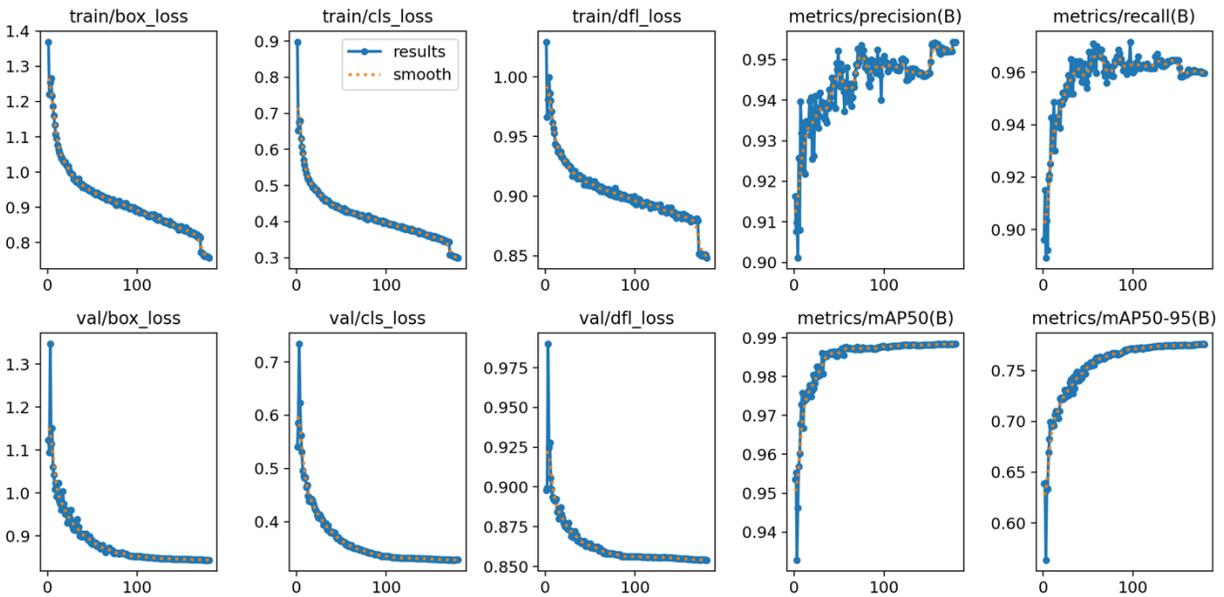

**Figure 4 YOLOv8s training and validation loss and accuracy curves for thermal deer detection**

Further, **Figure 5** illustrates the precision-confidence and recall-confidence curves, highlighting reliable calibration and high detection confidence for the model. The precision curve approaches 1.0 at a confidence threshold of 0.905, with a corresponding recall near 0.99 at low thresholds, underscoring the model's strong detection reliability in thermal imagery **(Figure 5a–b)**.

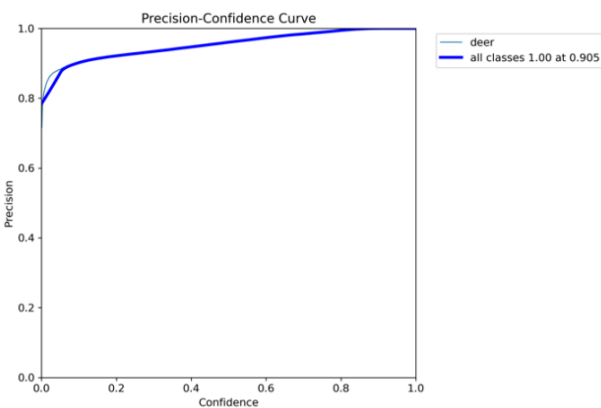
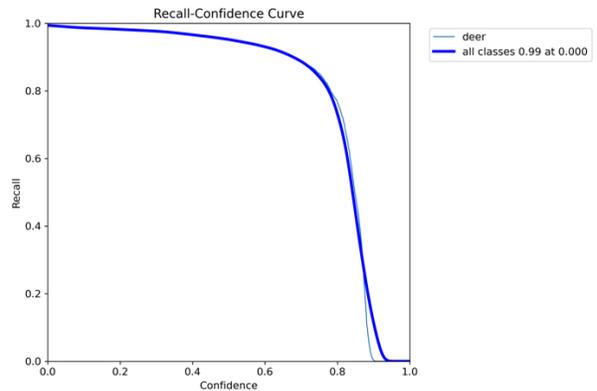

**Figure 5a Precision curve**     **Figure 5b Recall curve**

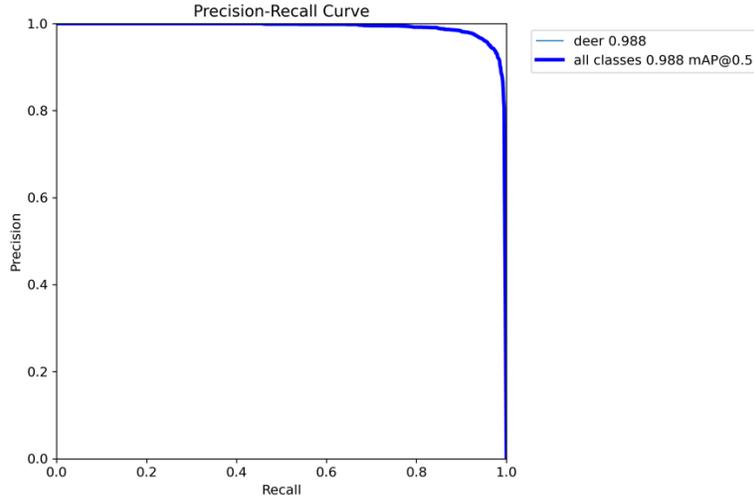

**Figure 6 Precision-recall curve**

   The overall model performance is demonstrated in **Figure 6**, which presents the precision-recall curve achieving a mean Average Precision (mAP@0.5) of 0.988. This high mAP value indicates strong model performance on the thermal deer detection task, with the precision-recall curve showing the trade-offs between detection accuracy and coverage across different confidence thresholds. The performance metrics suggest the model's potential applicability for deer detection in V2X safety systems, where both detection accuracy and false positive rates are important considerations.
   The model's classification performance is further illustrated by the confusion matrices in **Figures 7a and b**. **Figure 7b** presents the normalized confusion matrix, highlighting high true positive rates for the "deer" class and minimal misclassification between "deer" and "background." **Figure 7a** shows the absolute confusion matrix, detailing detection counts and confirming the model's robust separation between target and non-target classes.

**Table 2 YOLOv8s model performance evaluation**

| Performance Metric | Final Value | Baseline YOLOv8s/v11s | Previous Thermal Studies |
|---|---|---|---|
| mAP@0.5 | 98.84% | 47.0% (COCO) | 91.6% (Faster R-CNN) (3) |
| mAP@0.5:0.95 | 77.60% | ~50.0% | - |
| Precision | 95.44% | ~62% | - |
| Recall | 95.96% | ~59% | - |
| F1 Score | 95.70% | - | - |

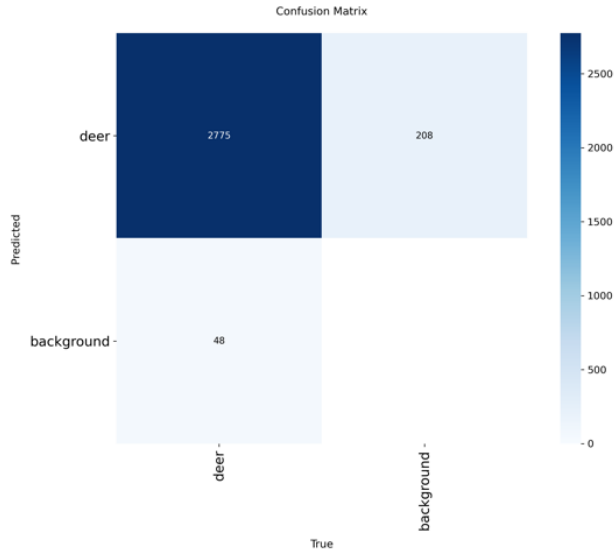 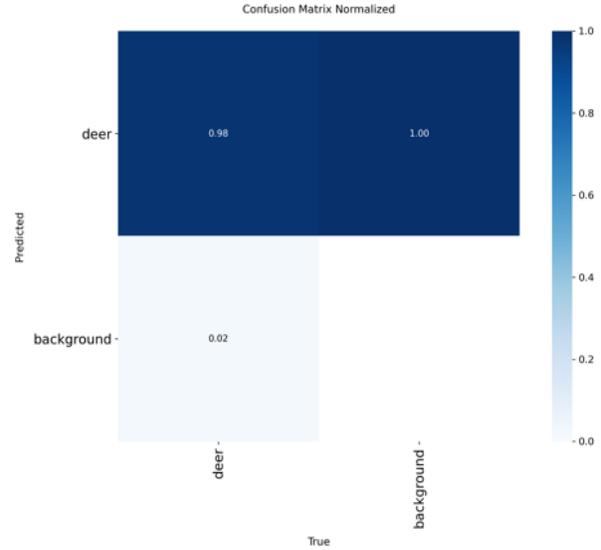

**Figure 7a. Confusion matrix for deer classification on the thermal image test set**

**Figure 7b. Normalized confusion matrix for classification on the thermal image test set**

**Field testing**
The system was field tested during a follow-up visit to Mars Hill, with trials conducted during evening and morning twilight periods, as well as at night. These conditions were intentionally selected to evaluate performance under low-visibility and challenging ambient environments. **Table 3** summarizes the system's accuracy at various distances from the thermal camera, highlighting both its practical deployment capabilities and its operational limitations. As detection range increases, performance declines, which is expected due to the resolution constraints of the thermal camera. At longer distances, the decrease in pixel density reduces feature clarity, making it more difficult to detect distant objects.

The field tests often involved scenarios in which deer were partially obscured by vegetation. This presented a realistic assessment of the detection system's robustness under real-world conditions. **Figure 8** illustrates that the system can reliably detect a fully visible deer positioned more than 50 feet away from the thermal camera with a class probability of 82%. **Figure 9** demonstrates the system's ability to detect a partially occluded deer at a similar distance, achieving a 72% class probability.

In each case, the detection pipeline encodes results into a standardized SDSM, which includes key positional and classification information. This message is transmitted in real time from the NVIDIA Jetson edge device to the Cohda MK6 OBU using UDP, and the Cohda OBU then broadcasts the detection message over the V2X network. **Figure 8** shows the thermal detection interface and the related SDSM data exchange, illustrating seamless integration between perception and V2X communication during deployment.

**Table 3 Field test evaluation based on camera detection range**

| Detection Range | Accuracy |
|---|---|
| <20 feet | 87-98% |
| 20-50 feet | 83-88% |
| 50-70 feet | 70-83% |
| 80-100 feet | 65-75% |
| >100 feet | variable |

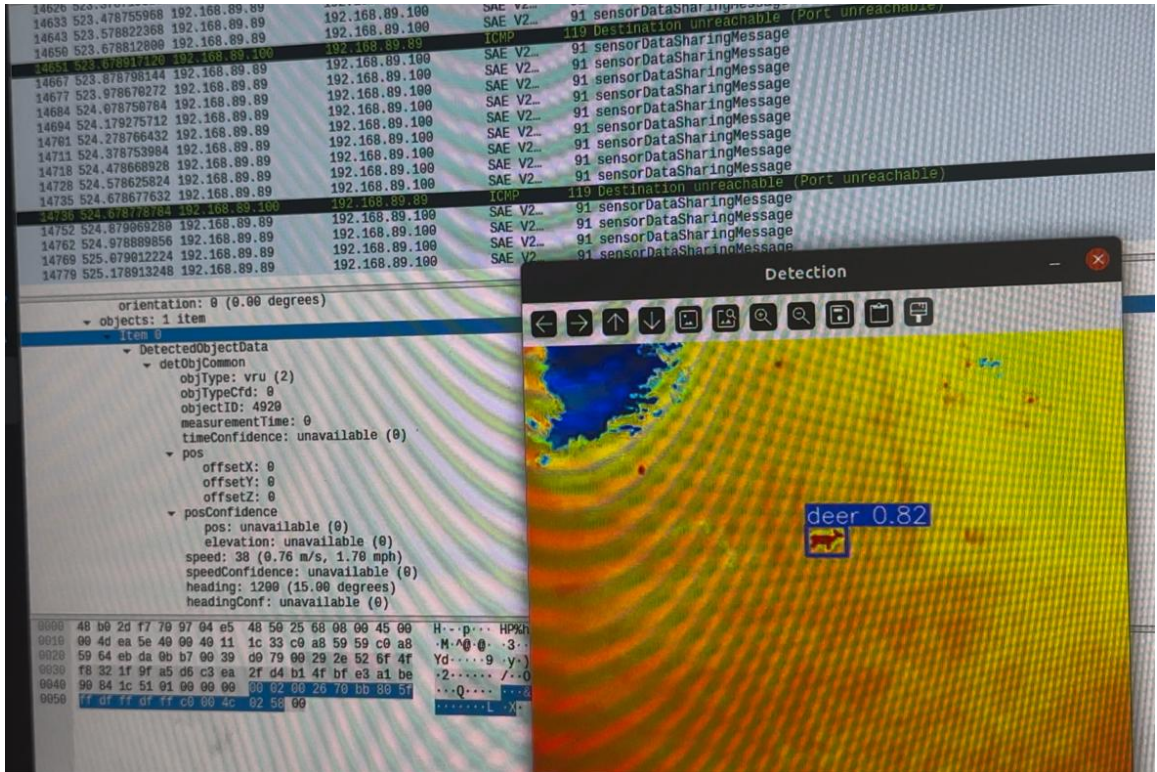

**Figure 8 Real-time thermal deer detection during field testing**

It is also notable that deer not detected when first entering the camera's field of view due to occlusion or distance are likely to be detected as they draw nearer and pixel density increases. When a deer is fully visible to the thermal camera, detection occurs reliably and with high probability, even at ranges approaching 100 feet. Overall, the consistent detection performance observed across a range of field conditions, along with real-time V2X message dissemination, confirms the system's practical value for wildlife-vehicle collision mitigation in operational settings.

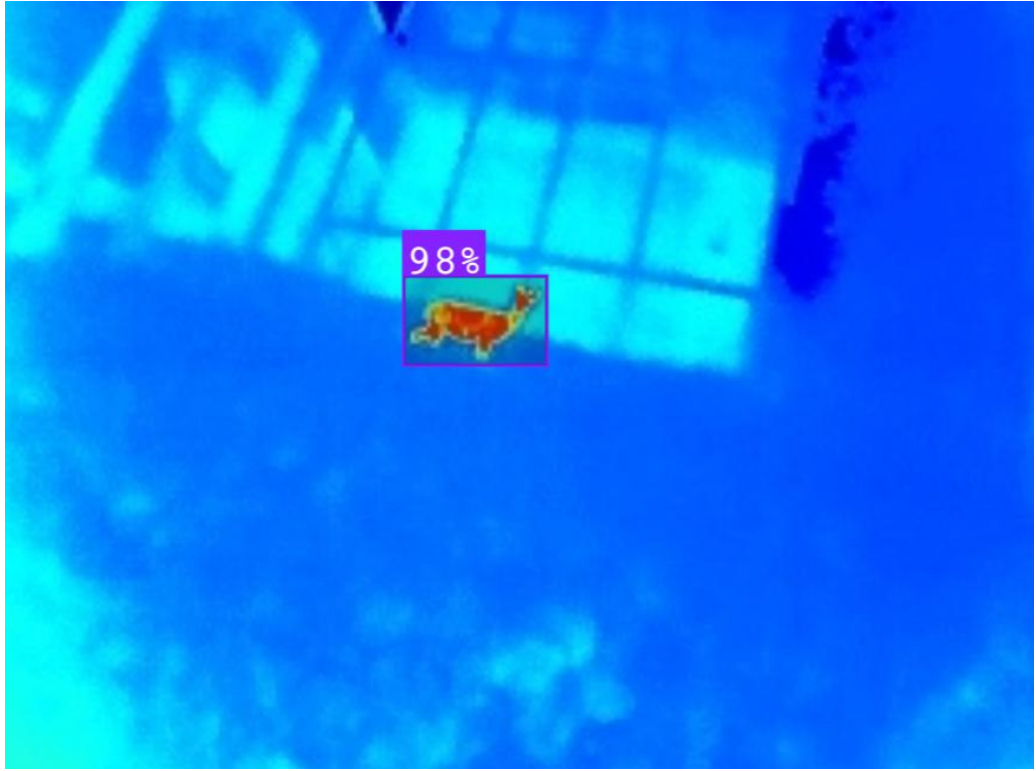

**Figure 9. Deer partially occluded by vegetation, illustrating detection challenges in real-world environments**

*Hardware performance characteristics and system latency analysis*
The Jetson Orin Nano, powered with a proper 15W supply, is capable of recording video from the TOPDON TC001 thermal camera at 25–30 frames per second over a 20-foot USB cable. When operating with the YOLOv8s deer detection, there is a drop in performance. The Jetson Orin Nano achieves 21-25 FPS inference under optimal conditions with a 1-foot USB cable connection on high power mode (15W). Performance sensitivity to cable length was observed, with the 20-foot cable reducing FPS to 5-7 during active detection and 15-18 FPS during idle periods, highlighting the importance of optimized hardware configurations. No other cable lengths were tested.

The results of latency tests are shown in **Table 4**. End-to-end system performance consistently meets automotive safety requirements with total latency under 100ms.

**Table 4 System latency**

| **Processing Stage** | **Latency** | **Maximum Latency** | **Critical Path Component** |
|---|---|---|---|
| Thermal Image Processing | 10ms | 25ms | Camera interface and preprocessing |
| YOLOv8s Inference | 40-50ms | 65ms | GPU computation |
| SDSM Generation | 8-10ms | 15ms | Message formatting |
| V2X Transmission | 10-15ms | 20ms | ASN.1 encoding and broadcast |
| Message Reception and Decoding | 10-15ms | 20ms | Adjacent vehicle processing |
| Alert Generation | 5-10ms | 15ms | Driver notification system |
| **Total System** | **<100ms** | **160ms** | **Complete detection-to-alert pipeline** |

*Environmental robustness assessment*
Under clear weather with moderate temperatures and little occlusion, the system achieved >95% detection accuracy within 40-foot range with false positive rates below 2% and missed detection rates under 4%. Nighttime, overcast daytime, fog and light rain had virtually no effect on the system. Conventional visible-light digital cameras under nighttime conditions achieved <60% accuracy, highlighting significant performance improvements provided by thermal sensing technology.

High ambient temperatures (>90°F) caused performance degradation to 75-80% accuracy due to reduced thermal contrast between deer and heated surfaces. Increased false positive rates (8-12%) occurred primarily from heated rock outcrops on the side of the road. These limitations can be mitigated through confidence threshold adjustment (>65%) and advanced filtering algorithms.

*V2X Communication performance validation*
V2X communication performance tests were conducted simultaneously with the deer detection testing. When a deer is detected over a series of frames, a warning message is broadcast via C-V2X to a nearby vehicle and an RSU. This was done regardless of the detection probability for test purposes. If the system was fully deployed, a detection probability threshold would be used to ensure that the warning message is credible.

**DISCUSSION AND LIMITATIONS**

This research demonstrates significant potential of a connected vehicle deer warning system through successful integration of thermal imaging, edge AI processing, and standards-compliant V2X messaging. The 98.84% mAP@0.5 achievement represents substantial improvement over baseline YOLOv8s performance (47% on COCO). The thermal imaging approach addresses fundamental limitations of visible-light systems, particularly during peak collision risk periods (dawn, dusk, and nighttime) when deer are most active with detection accuracy exceeding 95% regardless of ambient light. The sub-100ms end-to-end latency meets stringent automotive safety requirements while maintaining high detection reliability.

While the system evaluation provides proof of concept, a primary limitation of the system is the the detection range limitation. The effective detection range remains limited to approximately 100 feet due to thermal camera resolution constraints. The thermal camera used in this research is a low-cost low-resolution unit. A higher-resolution thermal sensor will increase detection range that would alert drivers sooner which is essential for mitigating high speed deer strikes.

Another limitation is high ambient temperature effects. The system performance degrades significantly in extreme heat conditions (>95°F) due to reduced thermal contrast between deer and surrounding heated surfaces. Road surfaces, rocks, and other objects absorbing solar energy create thermal noise, leading to increased false positive rates and occasional detection failures when deer are present on heated backgrounds. However, the primary need for a deer detection system is during periods where temperatures are cooler, and the sun is not overhead. Futher, false positives due to sun-heated objects typically occur with lower class probability scores (50-65%) enabling mitigation through threshold adjustment. Unfortunately, this threshold adjustment can impact distance detection.

False negative detection failures occur primarily when deer present rear-facing orientations without visible head/neck features, or when significantly occluded by vegetation. These scenarios highlight the importance of continuous monitoring and multiple detection opportunities as vehicles approach.

Another discussion point is that the current single-class detection limits system effectiveness to deer-only scenarios. Expanding to multi-class detection that includes other wildlife would enhance versatility but requires additional training data and computational resources.

**Economic and Deployment Considerations**
While thermal cameras and V2X equipment represent additional vehicle costs, potential savings from reduced collision damage, insurance claims, and human casualties provide compelling economic justification. With over $10 billion in annual DVC-related costs, even modest collision reduction rates would generate substantial economic benefits *(1)*. And wide scale deployment of thermal cameras and OBU hardware will reduce equipment costs.

One issue with the Cohda OBUs is that range is limited to about 1000 meters. Vehicle spacing can be far greater than 1000 meters in rural areas causing broadcasted warning messages to not reach other drivers. Thus, widespread deployment requires coordinated RSU installation for optimal message coverage and effectiveness. This will add significant infrastructure costs.

**CONCLUSION AND FUTURE WORK**
This research has presented a novel deer warning system designed to address a critical challenge mitigating deer-vehicle collisions through innovative integration of thermal imaging, edge artificial intelligence, and Vehicle-to-Everything communication technologies. The system demonstrates exceptional technical performance with 98.84% mAP@0.5 detection accuracy, sub-100ms end-to-end response latency, and 98% message delivery success within 1000m+ communication range.
A key innovation is the deep learning thermal sensing system capable of reliable deer detection during nighttime and low-visibility environmental conditions where conventional cameras fail. The real-time YOLOv8s inference on edge computing hardware achieves 21-25 FPS performance, and seamless integration with standards-compliant V2X messaging enabling collaborative safety awareness across connected vehicles and infrastructure.

Field validation in Mars Hill, North Carolina demonstrates practical effectiveness across diverse weather conditions and detection ranges. The thermal imaging approach provides accurate detection and driver warning capabilities during peak collision risk periods when deer are most active and traditional vision systems are least effective.

**Societal and Environmental Benefits**
Beyond immediate safety improvements, this technology may contribute to wildlife conservation by potentially reducing vehicle-related deer mortality. Economic implications include potential savings of billions of dollars annually through reduced collision damage, medical costs, and insurance claims.
The edge-based deployment approach reduces dependency on cellular network coverage, making advanced safety systems accessible in rural areas where traditional connected services may be unreliable, democratizing safety technology in areas with highest collision risk.

**Future Research Directions**
There is potential for significant future research to improve system performance, increase robustness and enhance versatility, they are discussed in the following paragraphs.

*Conduct tests with higher quality thermal sensors*
As stated previously, the thermal camera used in this research is a low-cost low-resolution unit. A higher-resolution thermal sensor with better clarity will increase detection range. A car traveling at 60 miles per hour will travel hundreds of feet before reacting to a deer that jumps in front of it. Thus, detection at significant distances is essential for reducing the most hazardous DVCs.

*Cloud and Cellular (CV2X) Integration*
Integrating cellular vehicle-to-everything (CV2X) technology and cloud-based infrastructure can extend the capabilities of the detection system well beyond the immediate V2X communication range. By using cellular networks, detection alerts and relevant data can be reliably transmitted to the cloud in real time. Looking ahead, we plan to further enhance this functionality by storing detection events in a centralized cloud database and integrating the information with mapping platforms such as Google Maps. This will allow drivers without onboard units (OBUs) to access timely wildlife hazard alerts directly through their navigation apps, ensuring that critical safety information reaches a broader range of road users and further reducing the risk of wildlife-vehicle collisions.

*Multi-Class Detection Enhancement*
Expanding detection capabilities to include multiple wildlife species (e.g. bear, moose, elk, caribou), domestic animals, and human pedestrians would significantly increase system versatility and safety impact. Transfer learning from existing thermal deer model weights could accelerate development while maintaining real-time performance requirements.

*Advanced Model Optimization*
Converting YOLOv8s to NVIDIA TensorRT format could provide additional inference acceleration, potentially enabling higher resolution processing or reduced latency. INT8 quantization techniques could reduce memory footprint and improve processing speed while maintaining acceptable accuracy levels.

*Predictive Behavior Integration*
Incorporating animal movement prediction based on historical tracking data could provide earlier warnings and more accurate collision risk assessment. Machine learning models trained on deer behavior patterns could enhance current detection-only approaches with predictive capabilities.

*Sensor Fusion Development*
Integrating thermal data with radar, LiDAR, or visible-light sensors could overcome single-modality limitations, potentially eliminating false positives from thermal mimics while improving performance in extreme weather conditions.

*Infrastructure Integration*
Enhanced RSU capabilities by incorporating other applications to provide mission critical information to drivers such as weather alerts, impending hazards, work zones ahead, etc.

**ACKNOWLEDGMENTS**
Funding for this research was provided by the U.S. Department of Transportation, Office of the Assistant Secretary for Research and Technology (OST-R), University Transportation Centers Program, through the Center for Regional and Rural Connected Communities (CR2C2) under Grant No. 69A3552348304. AI was used to check grammar, spell check, rephrase some sentences and to assist with literature search (Perplexity's GPT Model).

**AUTHOR CONTRIBUTIONS**
The authors confirm contribution to the paper as follows: study conception and design: W. Sarasua and H. Puppala; data collection: H. Puppala, S. Biyaguda, W. Sarasua, F. Farzinpour; analysis and interpretation of results: H. Puppala, W. Sarasua; draft M. Chowdhury manuscript preparation: H. Puppala, W. Sarasua, S. Biyaguda, F. Farzinpour, M. Chowdhury. Author. All authors reviewed the results and approved the final version of the manuscript.

**DECLARATION OF CONFLICTING INTERESTS**
The authors declared no potential conflicts of interest with respect to the research, authorship, and/or publication of this article.